\documentclass{article}

 \usepackage[preprint]{neurips_2026}

\usepackage[utf8]{inputenc}
\usepackage[T1]{fontenc}
\usepackage[english]{babel}
\usepackage{hyperref}       
\usepackage{url}  
\usepackage{microtype}
\usepackage{geometry}
\usepackage{amsmath, amssymb, amsthm, mathtools}
\usepackage{xcolor}
\usepackage{tcolorbox}
\tcbuselibrary{skins, breakable, theorems}
\usepackage{enumitem}
\usepackage{titlesec}
\usepackage{fancyhdr}

\newtcbtheorem[number within=section]{theorem}{Theorem}{
  enhanced,
  sharp corners,
  boxrule=0.5pt,
  colback=bgblue,
  colframe=themeblue,
  fonttitle=\bfseries,
  attach boxed title to top left={yshift=-2mm, xshift=3mm},
  boxed title style={sharp corners, size=small, colback=themeblue, colframe=themeblue, coltext=white},
  breakable,
  separator sign none
}{thm}

\newtcbtheorem[number within=section]{lemma}{Lemma}{
  enhanced,
  sharp corners,
  boxrule=0.5pt,
  colback=bggreen,
  colframe=themegreen,
  fonttitle=\bfseries,
  attach boxed title to top left={yshift=-2mm, xshift=3mm},
  boxed title style={sharp corners, size=small, colback=themegreen, colframe=themegreen, coltext=white},
  breakable,
  separator sign none
}{lem}

\newtcbtheorem[number within=section]{corollary}{Corollary}{
  enhanced,
  sharp corners,
  boxrule=0.5pt,
  colback=verylightred!30,   
  colframe=red!70,           
  fonttitle=\bfseries,
  attach boxed title to top left={yshift=-2mm, xshift=3mm},
  boxed title style={sharp corners, size=small, colback=red!70, colframe=red!70, coltext=white},
  breakable,
  separator sign none
}{cor}

\newtcolorbox{insightbox}{
  enhanced,
  sharp corners,
  boxrule=0.8pt,
  colback=white,
  colframe=themegray,
  fonttitle=\bfseries\color{themeblue},
  title=Core Insight,
  attach boxed title to top left={yshift=-2mm, xshift=3mm},
  boxed title style={sharp corners, size=small, colback=white, colframe=themegray, coltext=themeblue},
  breakable
}

\setlist{leftmargin=*, itemsep=0.3ex, topsep=0.5ex}

\usepackage{subfiles}

\usepackage{algorithm}
\usepackage{booktabs}      
\usepackage{xcolor}        
\usepackage{tabularx}      
\usepackage{colortbl}      
\usepackage{array}         
\usepackage{tabularx} 
\usepackage{rotating}
\usepackage{pdflscape}  
\usepackage{booktabs}   
\usepackage{algpseudocode}
\usepackage{thmtools}
\usepackage{wrapfig}

\declaretheorem[name=Proposition, numberwithin=section]{proposition}
\declaretheorem[name=Assumption, numberwithin=section]{assumption}
\declaretheorem[name=Remark, style=remark, numberwithin=section]{remark}

\title{Know When To Fold 'Em: Token-Efficient LLM Synthetic Data Generation via Multi-Stage In-Flight Rejection}

\author{%
  Anjir Ahmed Chowdhury\thanks{Use footnote for providing further information
    about author (webpage, alternative address)---\emph{not} for acknowledging
    funding agencies.} \\
  Department of Computer Science\\
  University of Houston\\
  \texttt{aachowd4@cougarnet.uh.edu} \\
  \And
  Syed Zawad \\
  IBM Research \\
  \texttt{szawad@ibm.com} \\
  \and
  Feng Yan \\
  Department of Computer Science\\
  University of Houston \\
  \texttt{fyan5@central.uh.edu} \\
}

\begin{document}

\maketitle

\begin{abstract}
While synthetic data generation with large language models (LLMs) is widely used in post-training pipelines, existing approaches typically generate full outputs before applying quality filters, leading to substantial token waste on samples that are ultimately discarded. To address this, we propose Multi-Stage In-Flight Rejection (MSIFR), a lightweight, training-free framework that detects and terminates low-quality generation trajectories at intermediate checkpoints before they reach full completion. MSIFR decomposes the generation process into sequential stages and applies fast rule-based validators to identify arithmetic inconsistencies, hallucination patterns, and formatting violations, enabling early rejection of faulty samples. We formalize in-flight rejection as a sequential decision process and show that any non-trivial discard policy reduces expected token consumption, with stage-wise savings increasing when rejection occurs earlier in the generation pipeline. We further demonstrate that conditional utility estimates form a martingale, ensuring that early, in-flight rejection does not bias the expected utility of retained samples. Across five instruction-tuned models and seven reasoning benchmarks, MSIFR reduces token consumption by 11\%–77\% as a standalone method, and up to 78.2\% when combined with early-exit methods, while preserving or improving evaluation accuracy. These results confirm that MSIFR provides a practical mechanism for improving the efficiency of LLM-based synthetic data generation without additional training or architectural changes.

\end{abstract}

\section{Introduction}
\begin{figure*}[!ht]
\centering
\includegraphics[width=0.99\textwidth]{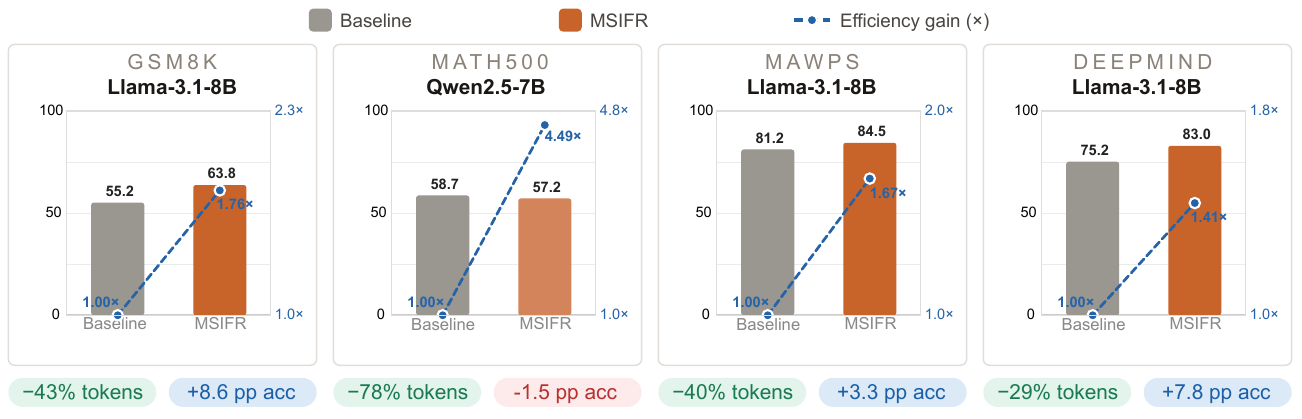}
\caption{MSIFR significantly reduces generation cost while maintaining competitive performance across benchmarks, achieving up to 78\% fewer tokens with only minor accuracy change, and in several cases improving accuracy by up to +8.6 percentage points over the baseline.}
\label{fig4}
\end{figure*}
The efficacy of large language models (LLMs) is fundamentally dependent on both the volume and 
quality of their training data, a relationship formalized by neural scaling laws 
\cite{kaplan2020scaling}. 
However, in the 
post-training domain encompassing supervised fine-tuning and reinforcement learning from human 
feedback (RLHF), the availability of high-quality human-annotated data is sparse due 
to privacy constraints, domain expertise requirements, and prohibitive annotation costs 
\cite{tang2025synthesizing,das2025revisiting,jalli2025synthetic}.

LLMs possess a unique dual capability in that they can both consume and produce human-like text and 
code, making them attractive as data generators for tasks where real data is costly or infeasible 
to obtain. Rather than manually labeling thousands of examples, one can prompt an LLM to 
synthesize diverse labeled instances at scale \cite{ieee2025synthetic,divekar2024synthesizrr,
alsakran2025llama}. As such, over the past several years, 
the LLM community has increasingly turned to LLM-generated content to augment training where 
real data is limited. 
Leading-edge models, including LLaMA \cite{dubey2024llama}, 
Falcon \cite{almazrouei2023falcon}, Qwen \cite{bai2023qwen}, and GPT-4 \cite{openai2024gpt4}, 
have all reported using synthetic data during post-training, with recent works further demonstrating their impact on model performance with as few as 20,000 instruction-response pairs \cite{tang2025synthesizing,tang2024synthesizing}, making them highly effective in low-resource settings, reducing annotation costs, and enabling data augmentation 
for improved robustness \cite{ieee2025synthetic,jalli2025synthetic,tang2025synthesizing}.

However, using LLMs as data generators introduces critical challenges in quality control and 
computational efficiency. Not all generated samples are equally valuable since flawed or low-quality 
problem-solution pairs can actively harm model performance by introducing noise into the training 
signal. Furthermore, LLM inference incurs computational cost proportional to the number of 
generated tokens. Continuing to elaborate on incoherent or incorrect samples wastes tokens that 
could be allocated to higher-quality generations, a strong concern given the already 
substantial computational budgets required for post-training. Early filtering of tokens by identifying and discarding weak samples at an early stage 
based on criteria such as logical inconsistency, arithmetic error, or formatting violation, a 
generation pipeline can prevent costly downstream expansion on unpromising trajectories. This 
reduces overall token consumption while improving the effective quality of training 
data. The model benefits from a cleaner learning signal, leading to better generalization and 
more stable optimization, while the system achieves a more favorable trade-off between 
computational cost and downstream performance. Existing early-exit methods ~\cite{akgul2025lynx,dai2025sgrpo,yang2025deer,laaouach2025halt} (e.g., S-GRPO achieves 40–61\% sequence-length reduction on single queries) operate on individual queries at inference time.

However, unlike prior works which shorten correct reasoning traces, we take a distinctively novel approach by \textit{identifying and terminating faulty trajectories before they incur full generation cost,} which saves significant resources compared to existing early-exit strategies. We propose \textbf{Multi-Stage In-Flight Rejection (MSIFR)}, a lightweight 
validation-driven framework for synthetic problem-solution generation that actively filters 
low-quality samples at both the problem stage and intermediate solution stages before full 
generation completes. In contrast to conventional approaches that generate entire solutions before 
applying any quality check, MSIFR decomposes generation into sequential stages and performs validation after each stage, checking for arithmetic consistency, hallucination indicators, 
and format correctness. If a sample fails any of these checks, generation is immediately 
terminated, preventing token expenditure on invalid trajectories. 

This design eliminates the need for complex reinforcement learning control while still 
leveraging structured decision points to prune poor generations early. By integrating 
validation-driven rejection criteria directly into the generation loop, MSIFR reduces unnecessary 
token usage, avoids propagating noisy or incorrect data, and ensures that only high-quality 
problem-solution pairs are fully generated and retained. Due to the completely separate strategy of our method, it can in fact act as a complementary framework to existing techniques, allowing for seamless integration and additive performance gains, which we demonstrate in our results.

We evaluate MSIFR across five instruction-tuned models, Qwen2.5-7B \cite{qwen2_5}, 
Meta Llama-3.1-8B \cite{llama3_1}, DeepSeek-7B \cite{deepseek}, 
Microsoft Phi-3-mini \cite{phi3}, and Mistral-7B \cite{mistral}, on seven 
benchmarks spanning mathematical reasoning and scientific knowledge: GSM8K \cite{gsm8k}, 
MATH500 \cite{math}, SVAMP \cite{svamp}, MAWPS 
\cite{mawps}, MathQA \cite{mathqa}, MMLU-Chem \cite{mmlu}, 
and the DeepMind Mathematics Dataset \cite{deepmind_math}. Compared with conventional 
full-generation pipelines, MSIFR reduces token consumption by 11\%–77\% as a standalone method (up to 42\% on GSM8K), and up to 78.2\% when combined with early-exit methods while preserving or 
improving evaluation accuracy across all benchmarks. Our contributions are:

\begin{itemize}
    \item We introduce MSIFR, a multi-stage validation framework that performs in-flight rejection during synthetic data generation, reducing token usage without requiring additional training.
    
    \item We provide a decision-theoretic formulation of early rejection and derive a stage-wise token savings decomposition, highlighting the importance of early-stage filtering.
    
    \item We show that conditional utility estimates form a martingale, ensuring that early rejection preserves unbiased expected utility.

    \item We empirically demonstrate consistent token savings and stable accuracy across multiple models and benchmarks, and show that MSIFR composes effectively with existing early-exit methods.
\end{itemize}
\section{Related Work}

We situate our work within two most directly related research threads: mid-generation 
filtering and dynamic abstention for unreliable trajectories, and synthetic data quality  control.

\subsection{Early Exit in Chain-of-Thought Reasoning}
A growing body of work addresses the overthinking problem, where reasoning 
models continue generating redundant tokens after arriving at a correct answer. \\
\textbf{Probe-based early exit.} LYNX \cite{akgul2025lynx} attaches exit decisions 
to naturally occurring reasoning cues, trains a lightweight probe on hidden states, 
and wraps scores in split conformal prediction for distribution-free control over 
premature exits. \\
\textbf{Reinforcement learning for early exit.} S-GRPO \cite{dai2025sgrpo} introduces 
group reward decay to encourage early termination, reducing sequence length by 
40-61\% on GSM8K, AIME 2024, and MATH-500 while simultaneously improving accuracy. 
DEER \cite{yang2025deer} uses entropy-based confidence at transition tokens to 
decide whether to stop. Think-or-Not \cite{yong2025think} performs tree-search over 
candidate continuations for reasoning mode selection. FlashThink \cite{jiang2025flashthink} 
queries verifiers on partial reasoning chains, trading off inference cost for 
earlier stopping decisions. \\
\textbf{PRM-guided early stopping.} ZGES \cite{vishwakarma2025prune} detects quality 
peaks in PRM-guided beam search using local reward z-scores, showing Pearson 
correlation > 0.91 with step quality. \\
\textbf{Training-free early exit.} HALT-CoT \cite{laaouach2025halt} computes Shannon 
entropy over the answer distribution after each reasoning step, halting when entropy 
falls below a threshold without training. \citet{wang2026detection} identify the 
Detection-Extraction Gap, where 52-88\% of CoT tokens are produced after the answer 
is already recoverable from a partial prefix, and exploit this asymmetry to truncate 
70-78\% of generation while improving accuracy by 1-5 percentage points. TERMINATOR \cite{nagle2025terminator} learns optimal stopping positions based on first answer token appearance.

\subsection{Mid-Generation Filtering and Dynamic Abstention}

The most related line of work addresses terminating \textbf{unpromising 
trajectories} that are unlikely to yield correct answers regardless of 
generation, a setting that closely mirrors our problem formulation. \\
\textbf{Early rejection with learned rewards.} \citet{khan2025earlyrejection} apply 
PRMs mid-generation to reject suboptimal candidates before a full reasoning step 
completes, proving that the risk of discarding optimal beams decreases exponentially 
with generation length. \\
\textbf{Confidence-aware mid-generation filtering.} DeepConf \cite{deepconf2025} 
combines parallel thinking with confidence-aware filtering using model-internal token 
distributions to discard low-quality reasoning traces during or after generation, 
achieving up to 84.7\% token reduction on AIME 2025 without additional training. 
Unlike MSIFR, DeepConf operates at inference time on a single reasoning task and relies 
on learned confidence signals rather than lightweight rule-based validators. \\
\textbf{Principled abstention.} \citet{nachshon2026knowing} formalize dynamic 
abstention within a KL-regularized RL framework, proving that value-thresholding 
strictly dominates fixed-position abstention and achieving up to 91\% selective 
accuracy at 90\% abstention rates on OlympiadBench. \citet{liao2025lost} show that 
first-step errors disproportionately degrade final answer quality and propose 
reward model-based first-step selection achieving up to 70\% inference cost reduction 
without accuracy loss. Both works target inference-time reasoning and require trained 
reward components, whereas MSIFR is fully training-free and operates at the level of 
dataset construction. \\
\textbf{Clarification-Aware Abstention.}
Abstain-R1 \cite{zhai2026abstain} studies calibrated abstention for unanswerable 
queries using a clarification-aware RLVR reward, achieving behavior competitive with 
much larger systems including DeepSeek-R1 on Abstain-Test, Abstain-QA, and SelfAware. 
This work addresses a complementary problem of abstaining on unanswerable inputs 
rather than filtering low-quality generative trajectories.

\subsection{Positioning MSIFR}

Early-exit methods optimize the termination of correct but verbose trajectories at inference time and therefore do not address the problem of faulty data generation. Early rejection approaches~\cite{khan2025earlyrejection} and DeepConf~\cite{deepconf2025} discard unpromising trajectories mid-generation, but rely on trained model components and are designed for single-query inference rather than large-scale dataset construction. Dynamic abstention~\cite{nachshon2026knowing} provides strong theoretical foundations for stopping decisions, but has not been instantiated as a lightweight, training-free, online validation policy. Synthetic data pipelines~\cite{tang2025synthesizing,zheng2023judging} typically apply quality filtering only post-hoc, after the full generation cost has already been incurred.

In contrast, our goal is to minimize token waste from faulty synthetic generations by detecting and rejecting them before full completion. MSIFR operates in a training-free, dataset construction setting, where many samples are generated and low-quality trajectories must be eliminated efficiently. It introduces a multi-stage, in-flight rejection mechanism that validates partial outputs during generation and immediately terminates faulty trajectories, thereby avoiding unnecessary token expenditure without waiting until the end of generation to reject a faulty sample.

\section{MSIFR Framework}
\label{sec:theory}

\begin{figure*}[!ht]
\centering
\includegraphics[width=0.9\textwidth]{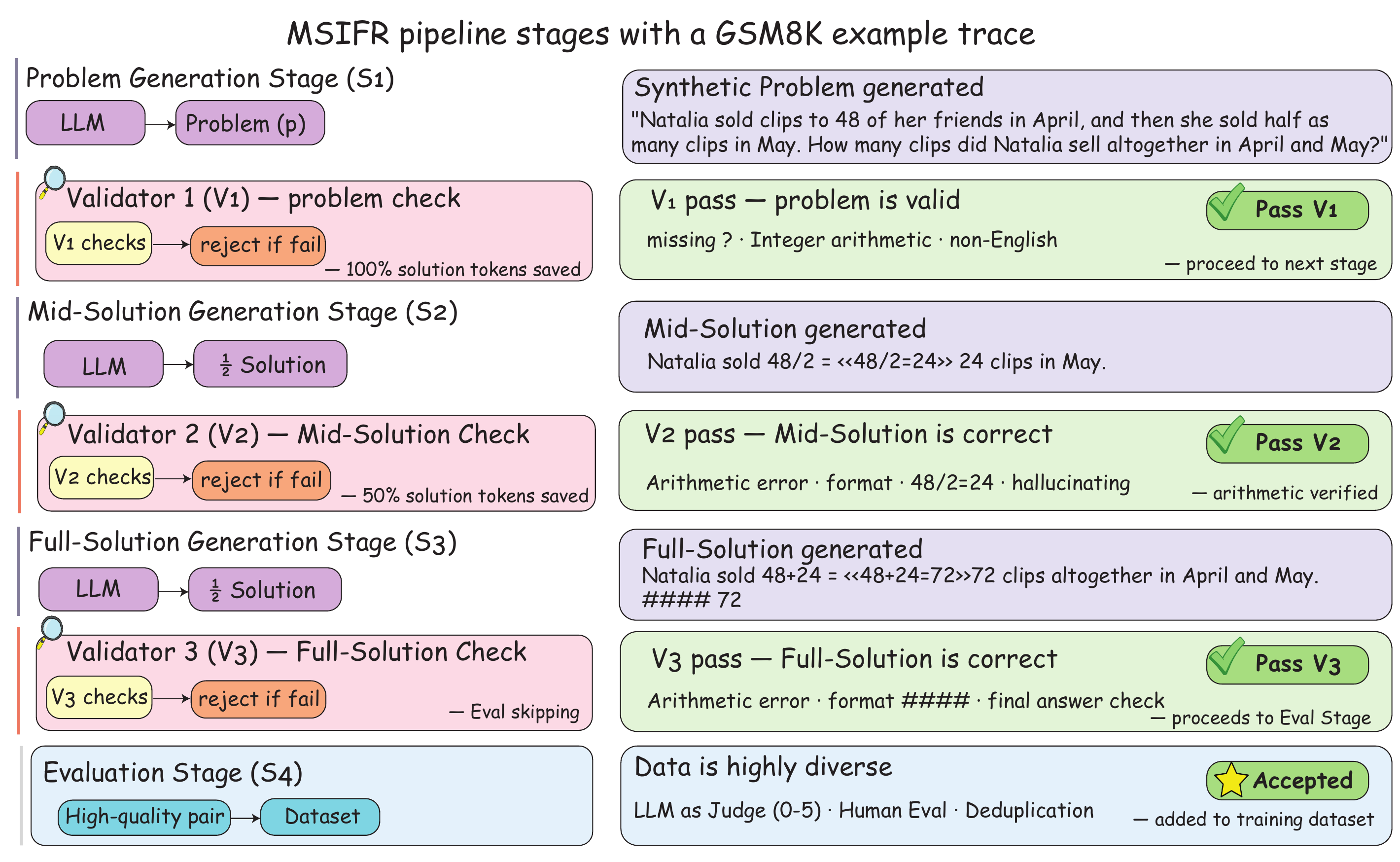}
\caption{MSIFR pipeline for synthetic data generation. Validators 
$(V_1$--$V_3)$ filter low-quality partial generations at each stage. The three validator stages (V1–V3) correspond to the Well-Posedness Enforcer (WPE), Reasoning Trace Auditor (RTA), and Solution Convergence Validator (SCV) described in Appendix \ref{app:validator}. The GSM8k trace 
illustrates how arithmetic or formatting errors trigger early rejection, preventing 
unnecessary token expenditure before final LLM and human evaluation.}
\label{fig1}
\end{figure*}

Figure~\ref{fig1} illustrates the \textsc{MSIFR} pipeline.
Given an input prompt, generation proceeds through four stages:
problem generation (\textbf{S1}), mid-solution generation (\textbf{S2}),
full solution generation (\textbf{S3}), and final evaluation (\textbf{S4}).
At each of the first three stages, a lightweight validator~$V_t$ inspects the
partial output and immediately discards any trajectory that fails.
Only samples passing all intermediate checks proceed to S4, where
LLM judging, human verification, and deduplication determine dataset inclusion.
Algorithm~\ref{alg:MSIFR} formalises this procedure.
We now derive the efficiency guarantees underpinning \textsc{MSIFR}.

\subsection{Problem Setup}
\label{sec:setup}

We model synthetic data generation as a sequential stochastic process
with $T = 4$ stages.
Given input $x$, the LLM produces a chain of partial outputs following standard autoregressive language model generation formulations~\cite{brown2020,openai2024gpt4}.
\begin{equation}
  y^{(t)} \sim p_\theta\!\left(\,\cdot \mid y^{(t-1)},\, x\right),
  \qquad y^{(0)} = \varnothing,
  \label{eq:gen}
\end{equation}
where $p_\theta$ is the model's conditional distribution. 
Each stage incurs an incremental token cost $\Delta c_t > 0$, with
$\Delta c_1 \ll \Delta c_2 \ll \Delta c_3 \ll \Delta c_4$, so the
total cost of unconstrained generation is
\begin{equation}
  C_{\mathrm{full}} \;=\; \sum_{t=1}^{T} \Delta c_t.
\end{equation}
\textbf{Remark:} Because $\Delta c_{1} \approx 0$ relative to the later solution stages, the effective token savings are primarily driven by early rejection at $S_{2}$ ($\Delta c_{2}$) and $S_{3}$ ($\Delta c_{3}$). The lower bound in Proposition~3.1,
$\Delta c_{T}\cdot \Pr(\tau < T)$, is tight only when rejection is concentrated near the final stage. In contrast, rejection occurring earlier at $S_{2}$ or $S_{3}$ produces super-linear savings relative to this bound, consistent with the empirical results reported in Table~\ref{tab:main_results}.
A completed sample $y = y^{(T)}$ is evaluated by a set of
binary validators $\mathcal{V} = \{v_1, \dots, v_m\}$,
$v_i(y) \in \{0,1\}$, with composite utility
\begin{equation}
  q(y) \;=\; \prod_{i=1}^{m} v_i(y).
\end{equation}
If generation is halted at stage $\tau < T$, we set $q(y) = 0$,
ensuring that only fully generated and validated samples contribute
positive utility.

\subsection{Decision-Theoretic Objective}
\label{sec:objective}

A \emph{discard policy} $\pi$ inspects the partial output at each
stage and decides whether to continue or abort.
Stopping at stage~$\tau$ incurs cumulative cost
$C(\tau) = \sum_{t=1}^{\tau} \Delta c_t$.
The policy optimises
\begin{equation}
  J(\pi)
  \;=\;
  \mathbb{E}_\pi\!\left[
    q\!\left(y^{(T)}\right) \cdot \mathbf{1}_{\{\tau = T\}}
    \;-\; \lambda\, C(\tau)
  \right],
  \label{eq:objective}
\end{equation}
where $\lambda > 0$ governs the utility cost trade-off. This objective is conceptually related to KL-regularized sequence optimization and abstention-based stopping formulations studied in prior work~\cite{jaques2017,ziegler2019fine,nachshon2026knowing}.
The Bayes optimal policy at stage~$t$ compares
$\mathbb{E}\!\left[q(y^{(T)}) \mid y^{(t)}\right]$ against the
expected future token cost since this conditional expectation is
intractable to compute online, we introduce a tractable surrogate
in Section~\ref{sec:surrogate}.

\subsection{Surrogate Utility Estimation}
\label{sec:surrogate}

At stage~$t$ we apply $K_t$ lightweight rule-based checks and define
$s_t(y^{(t)}) \in \{0, \dots, K_t\}$ as the number of checks passed.
The surrogate relies on the following mild ordinal assumption.

\begin{assumption}[Monotone score--quality correlation]
\label{ass:monotone}
For each stage $t$, there exists a non-decreasing function
$\phi_t\colon \{0,\dots,K_t\} \to [0,1]$ such that
\begin{equation}
  \mathbb{E}\!\left[q\!\left(y^{(T)}\right) \mid y^{(t)}\right]
  \;=\;
  \phi_t\!\left(s_t\!\left(y^{(t)}\right)\right) + \epsilon_t,
  \label{eq:surrogate}
\end{equation}
where $\epsilon_t$ is zero-mean noise independent of $s_t$.
\end{assumption}

Assumption~\ref{ass:monotone} requires only ordinal consistency between the validation score and expected final quality, rather than calibrated probability estimation. This assumption is considerably weaker than the calibration assumptions commonly used in conformal prediction and uncertainty quantification~\cite{vovk2005,angelopoulos2021}.
and is verified empirically in Section~\ref{results}.

\subsection{Token-Efficiency Guarantee}
\label{sec:token-guarantee}

\begin{proposition}[Strict token savings]
\label{prop:token}
If\/ $\mathbb{P}(\tau < T) > 0$ and $\Delta c_t > 0$ for all~$t$, then
\begin{equation}
  \mathbb{E}[C(\tau)] \;<\; C_{\mathrm{full}}.
  \label{eq:savings}
\end{equation}
Moreover, the expected savings admit the decomposition
\begin{equation}
  C_{\mathrm{full}} - \mathbb{E}[C(\tau)]
  \;=\;
  \sum_{t=1}^{T} \Delta c_t \cdot \mathbb{P}(\tau < t)
  \;\geq\;
  \Delta c_T \cdot \mathbb{P}(\tau < T)
  \;>\; 0.
  \label{eq:decomp}
\end{equation}
\end{proposition}

\paragraph{Proof sketch.}
Writing the expected cost as
$\mathbb{E}[C(\tau)] = \sum_{t=1}^{T} \Delta c_t\, \mathbb{P}(\tau \geq t)$
and using $\mathbb{P}(\tau \geq T) < 1$ (which follows from
$\mathbb{P}(\tau < T) > 0$) yields~\eqref{eq:savings}.
Subtracting and applying the identity
$\mathbb{P}(\tau < t) = 1 - \mathbb{P}(\tau \geq t)$ gives the
decomposition in~\eqref{eq:decomp}.
The lower bound follows because $\mathbb{P}(\tau < t)$ is
non-decreasing in~$t$ with $\mathbb{P}(\tau < 1) = 0$,
so the dominant term appears at $t = T$. \hfill$\square$

\smallskip
Equation~\eqref{eq:decomp} has a natural interpretation.
Each stage~$t$ contributes savings proportional to the probability
that a trajectory is already discarded before reaching it.
Because costs are ordered $\Delta c_1 \ll \cdots \ll \Delta c_T$,
early rejection of low-quality trajectories yields
super-linear savings relative to a late-stage filter.

\subsection{Unbiasedness of In-Flight Rejection}
\label{sec:martingale}

A natural concern is whether early stopping introduces a systematic
bias into the utility of accepted samples.
The following result shows that this cannot occur.

\begin{proposition}[Martingale unbiasedness]
\label{prop:martingale}
Let $\mathcal{F}_t = \sigma\!\left(y^{(1)}, \dots, y^{(t)}\right)$ and
define $M_t = \mathbb{E}\!\left[q(y^{(T)}) \mid \mathcal{F}_t\right]$.
Then $\{M_t\}_{t=0}^{T}$ is a uniformly integrable martingale, and
for any stopping time $\tau \leq T$,
\begin{equation}
  \mathbb{E}[M_\tau] \;=\; \mathbb{E}\!\left[q\!\left(y^{(T)}\right)\right].
  \label{eq:unbiased}
\end{equation}
\end{proposition}

\paragraph{Proof sketch.}
The martingale property
$\mathbb{E}[M_{t+1} \mid \mathcal{F}_t] = M_t$
follows by iterated conditioning on $\mathcal{F}_t \subseteq \mathcal{F}_{t+1}$.
Uniform integrability holds because $q\!\left(y^{(T)}\right)\in\{0,1\}$ is bounded. The discard policy $\pi$ uses only information available in the filtration
\[
\mathcal{F}_{t}=\sigma\!\left(y^{(1)},\ldots,y^{(t)}\right)
\]
at stage $t$, so the stopping variable $\tau$ is measurable with respect to $\{\mathcal{F}_{t}\}$ and therefore constitutes a valid stopping time. Equation~\eqref{eq:unbiased} then follows from the Optional Stopping Theorem~\citep{williams1991}, which applies since $\tau \leq T < \infty$.
\hfill$\square$

\smallskip
Proposition~\ref{prop:martingale} guarantees that the expected quality
of a sample surviving to any intermediate stage equals the expected quality
under full generation.
The rejection rule therefore cannot inflate or deflate the estimated
utility of retained trajectories, a property critical to the reliability
of the downstream training signal.

\subsection{Algorithm}
\label{sec:algorithm}

\begin{wrapfigure}{r}{0.44\textwidth}
  \vspace{-25pt} 
  \begin{minipage}{0.44\textwidth}
    \begin{algorithm}[H]
      \caption{Multi-Stage In-Flight Rejection (\textsc{MSIFR})}
      \label{alg:MSIFR}
      \scriptsize
      \begin{algorithmic}[1]
        \Require Input $x$; thresholds $\{\lambda_t\}_{t=1}^{T-1}$;
                 validators $\{f_t\}_{t=1}^{T-1}$; final checks $\mathcal{V}$
        \State $y^{(0)} \gets \varnothing$
        \For{$t = 1$ to $T-1$}
          \State Generate $y^{(t)} \sim p_\theta(\cdot \mid y^{(t-1)}, x)$
          \If{$f_t(y^{(t)}) \leq \lambda_t$}
            \State \Return $\varnothing$ \Comment{In-Flight Rejection}
          \EndIf
        \EndFor
        \State Generate $y^{(T)} \sim p_\theta(\cdot \mid y^{(T-1)}, x)$
        \If{$\prod_i v_i(y^{(T)}) = 1$}
          \State \Return $y^{(T)}$
        \Else
          \State \Return $\varnothing$
        \EndIf
      \end{algorithmic}
    \end{algorithm}
  \end{minipage}
  \vspace{-15pt}    
\end{wrapfigure}

Algorithm~\ref{alg:MSIFR} implements \textsc{MSIFR}.
At each stage~$t$, Generation proceeds only when the validation score $f_t\!\left(y^{(t)}\right)$ strictly exceeds the stage-specific threshold $\lambda_t$, i.e., $f_t\!\left(y^{(t)}\right) > \lambda_t .$ Otherwise, the trajectory is immediately rejected, as specified in lines 4--5 of Algorithm~\ref{alg:MSIFR}. In practice, MSIFR implements $T=4$ stages, where the first three correspond to intermediate validation checkpoints and the final stage performs full solution verification.
If the score falls at or below~$\lambda_t$, the trajectory is
immediately abandoned and no further tokens are consumed.
Only trajectories passing all $T{-}1$ intermediate checks proceed
to full generation; the completed sample is then accepted if and
only if every final validator $v_i$ returns~$1$.
This design ensures that the total token cost is incurred
exclusively for high-confidence trajectories, directly realising
the savings guaranteed by Proposition~\ref{prop:token}.

\subsection{Relationship to Early-Exit Methods}
\label{sec:early-exit}

Early-exit methods~\cite{akgul2025lynx,dai2025sgrpo,yang2025deer,laaouach2025halt}
reduce computation \emph{per token} by terminating the forward pass
at shallower layers once the model is sufficiently confident.
\textsc{MSIFR} reduces the total \emph{number of tokens} by aborting
entire low-quality generation trajectories before they complete.
These two mechanisms operate on orthogonal cost axes and are composable.

\begin{remark}[Composability]
\label{rem:composable}
Early exit reduces FLOPs per token; in-flight rejection reduces
tokens per sample.
Applying both simultaneously yields compounding efficiency gains:
Table~\ref{tab:method_comparison_minimal} shows that
\textsc{MSIFR}$+$\textsc{LYNX} achieves $5{\times}$ throughput
over conventional generation with no accuracy penalty.
\end{remark}

\begin{figure*}[!ht]
\centering
\includegraphics[width=0.9\textwidth]{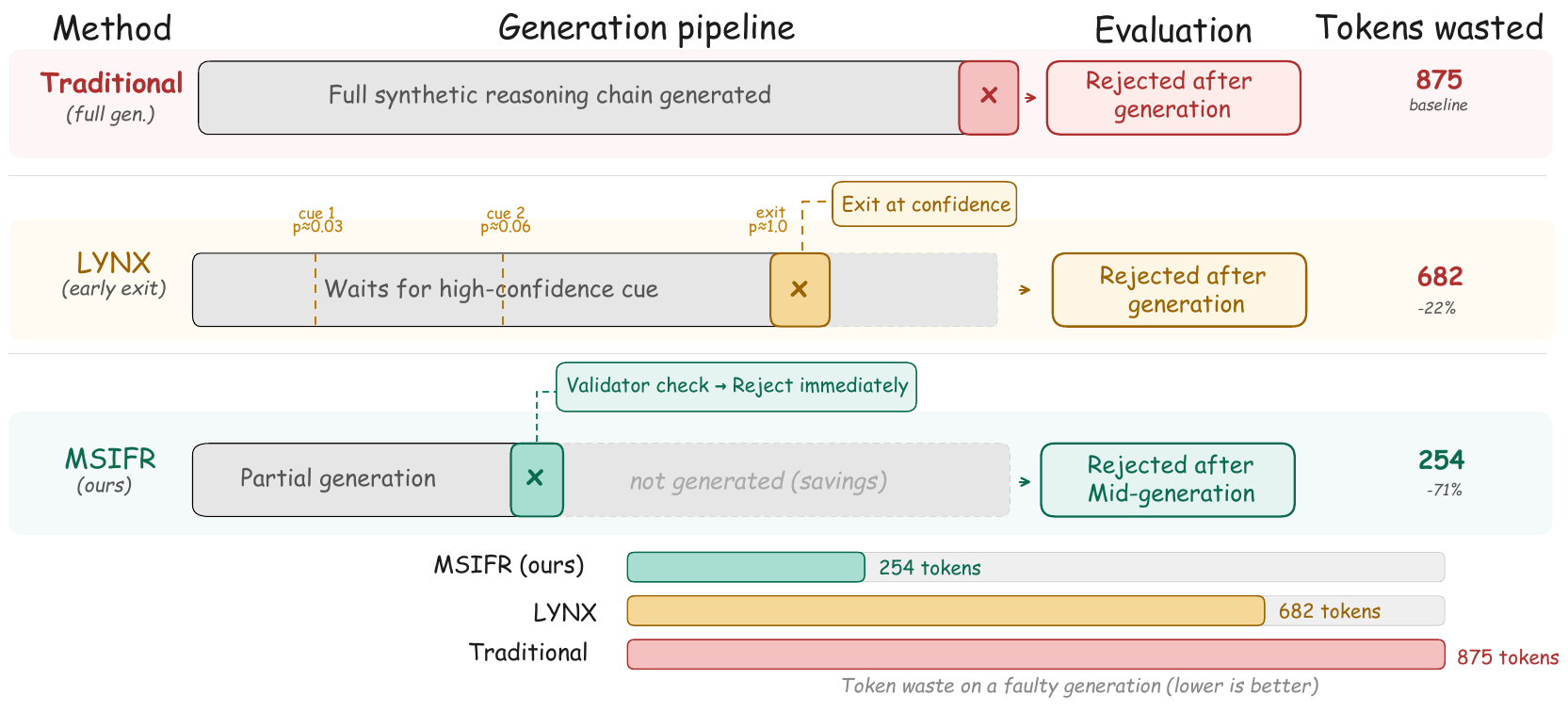}
\caption{Pipeline comparison of rejection strategies for synthetic reasoning data generation. Given a multi-step arithmetic problem (Figure~\ref{fig2}), all three methods ultimately reject the same incorrectly reasoned solution, where the error originates from computing $3 \times 60$ instead of $4 \times 60$. The Traditional baseline performs validation only after full generation, resulting in 875 wasted tokens. LYNX reduces token usage to 682 ($-22\%$) by applying probabilistic early-exit cues; however, its stopping decision is triggered only after sufficient confidence is accumulated, typically late in the generation process. In contrast, MSIFR performs in-flight rejection by validating intermediate outputs and terminates the faulty trajectory as soon as the inconsistency is detected, at 254 tokens ($-71\%$ relative to the baseline). This comparison highlights that early validation enables substantially greater token savings by preventing unnecessary continuation of incorrect reasoning.}
\label{fig3}
\end{figure*}

\section{Results}
\label{results}
\subsection{Models \& Dataset}

We evaluate MSIFR on five open-source instruction-tuned language models: \textbf{Qwen2.5-7B-Instruct} \citep{qwen2_5}, \textbf{Llama-3.1-8B-Instruct} \citep{llama3_1}, \textbf{DeepSeek-LLM-7B-Chat} \citep{deepseek}, \textbf{Phi-3-mini-4k-instruct} \citep{phi3}, and \textbf{Mistral-7B-Instruct-v0.3} \citep{mistral}. All models are evaluated across seven mathematical reasoning benchmarks: \textbf{GSM8K} \citep{gsm8k}, \textbf{MATH500} (a 500-problem subset of MATH \citep{math}), \textbf{MMLU-Chem} (chemistry subset of MMLU \citep{mmlu}), \textbf{SVAMP} \citep{svamp}, \textbf{MAWPS} \citep{mawps}, \textbf{MathQA} \citep{mathqa}, and the \textbf{DeepMind Mathematics Dataset} \citep{deepmind_math}. For automatic evaluation of generation quality, we employ \textbf{Llama-3-13B-Instruct} \citep{llama3} as a judge model, which scores completions for correctness and coherence, following prior work on LLM-as-a-judge evaluation \citep{zheng2023judging}.

\subsection{Experiment Setup}

All experiments are conducted on a system equipped with two NVIDIA RTX 4090 GPUs (24GB VRAM each). We utilize \textbf{vLLM} \citep{kwon2023vllm} for efficient batched inference, enabling high-throughput generation through PagedAttention and continuous batching. The generation configuration uses temperature ($T_p$ = 0.7) for problem generation to encourage diversity and ($T_s$ = 0.0) for solution generation to ensure deterministic reasoning. We set the batch size to 64 samples per generation step. Generated data is stratified into three difficulty tiers based on a scalar score (d): easy ((d $\in$ [1, 50))), medium ((d $\in$ [50, 500))), and hard ((d $\in$ [500, 2000])). We adopt a hybrid sampling strategy consisting of 30\% few-shot prompting and 70\% LLM-only prompting. The early-discard pipeline consists of four validator stages: Before Solution, Mid Solution, Full Solution, and After Evaluation, with detailed validator functions and chat templates provided in the Appendix. For quality assessment, we use \textbf{Llama-3-13B-Instruct}-as-a-judge scoring on a 1–5 scale, where scores are thresholded for acceptance (score $\leq$ 3). All results are reported with three decimal precision for accuracy and token metrics, and throughput is measured in problem solution pairs per hour. To ensure reproducibility, all experiments are conducted with the same random seed (seed = 42).

\subsection{Results Across Benchmarks}
Table~\ref{tab:main_results} presents total token consumption, evaluation accuracy, and throughput of five LLM models under our multi-stage in-flight rejection (MSIFR) framework across seven benchmarks. The Traditional baseline token counts are computed by fully generating all samples without early rejection, ensuring a consistent comparison against MSIFR under identical generation settings. Across all models and benchmarks, MSIFR consistently reduces total token consumption relative to this baseline by identifying and terminating faulty generations before completion, recovering compute that would otherwise be wasted on low-quality sequences.
Llama-3.1-8B achieves the strongest overall efficiency profile, attaining the lowest total token count on four of seven benchmarks (GSM8K: 7.46M, MMLU-Chem: 0.211M, MAWPS: 1.589M, DeepMind: 10.726M) while simultaneously achieving the highest evaluation accuracy on four benchmarks (MATH500: 0.619, MMLU-Chem: 0.443, MAWPS: 0.845, DeepMind: 0.830). Token savings over the Traditional baseline range from approximately 11\% to 53\% depending on the model and benchmark, confirming that catching faulty generations mid-flight substantially reduces wasted computation without modifying the underlying generator. Notably, accuracy under in-flight rejection meets or exceeds the Traditional baseline in the majority of model–benchmark pairs, indicating that terminating low-confidence sequences early concentrates accepted samples among higher-quality continuations.

\begin{table}[!ht]
\centering
\scriptsize
\caption{Comparing performance of MSIFR across seven benchmarks. All models target fixed accepted samples per benchmark. Lower Total Token ($\downarrow$) and higher Eval Accuracy ($\uparrow$) are better. $\dagger$ Traditional = full generation without early discard.}
\label{tab:main_results}
\begin{tabular}{l l c c c c c c c}
\toprule
\textbf{Model} & \textbf{Metric} & \textbf{GSM8K} & \textbf{MATH500} & \textbf{MMLU-Chem} & \textbf{SVAMP} & \textbf{MAWPS} & \textbf{MathQA} & \textbf{DeepMind} \\
\midrule
Traditional$\dagger$ & Total Token $\downarrow$ & 13.167M & 14.526M & 0.293M & 3.363M & 2.653M & 18.159M & 15.148M \\
& Eval Acc $\uparrow$ & 0.552 & 0.587 & 0.421 & 0.780 & 0.812 & 0.652 & 0.752 \\
& Throughput $\uparrow$ & 20,369 & 10,157 & 15,763 & 16,324 & 18,429 & 16,114 & 17,369 \\
\midrule
Qwen2.5-7B & Total Token $\downarrow$ & 9.371M & 3.232M & 0.243M & 1.980M & 1.611M & 11.413M & 10.919M \\
& Eval Acc $\uparrow$ & 0.580 & 0.572 & 0.434 & 0.832 & 0.836 & 0.833 & 0.815 \\
& Throughput $\uparrow$ & 32,141 & 11,247 & 15,076 & 15,634 & 19,344 & 16,286 & 17,884 \\
\midrule
Llama-3.1-8B & Total Token $\downarrow$ & 7.462M & 4.353M & 0.211M & 1.950M & 1.589M & 13.679M & 10.726M \\
& Eval Acc $\uparrow$ & 0.638 & 0.619 & 0.443 & 0.803 & 0.845 & 0.695 & 0.830 \\
& Throughput $\uparrow$ & 31,356 & 9,950 & 16,141 & 16,268 & 20,126 & 17,530 & 18,199 \\
\midrule
DeepSeek-7B & Total Token $\downarrow$ & 11.585M & 7.894M & 0.289M & 2.730M & 2.012M & 15.482M & 13.891M \\
& Eval Acc $\uparrow$ & 0.501 & 0.558 & 0.398 & 0.640 & 0.710 & 0.514 & 0.643 \\
& Throughput $\uparrow$ & 28,038 & 8,549 & 13,174 & 14,443 & 17,874 & 19,045 & 18,012 \\
\midrule
Phi-3-mini & Total Token $\downarrow$ & 7.485M & 5.025M & 0.229M & 2.299M & 1.744M & 12.769M & 11.909M \\
& Eval Acc $\uparrow$ & 0.603 & 0.529 & 0.413 & 0.760 & 0.820 & 0.744 & 0.755 \\
& Throughput $\uparrow$ & 31,856 & 8,735 & 15,927 & 14,612 & 18,789 & 16,668 & 17,412 \\
\midrule
Mistral-7B & Total Token $\downarrow$ & 9.826M & 10.274M & 0.246M & 2.773M & 2.133M & 16.935M & 14.014M \\
& Eval Acc $\uparrow$ & 0.532 & 0.512 & 0.401 & 0.630 & 0.670 & 0.561 & 0.630 \\
& Throughput $\uparrow$ & 27,406 & 8,872 & 14,224 & 14,035 & 17,618 & 19,035 & 18,000 \\
\bottomrule
\end{tabular}
\end{table}

Qwen2.5-7B achieves best-in-class accuracy on SVAMP (0.832) and MathQA (0.833) with moderate token savings, suggesting that a stronger reasoning prior raises the quality floor of accepted samples under the same rejection budget. DeepSeek-7B and Mistral-7B exhibit the weakest efficiency gains: both consume more total tokens than most alternatives and achieve the lowest accuracy on the majority of benchmarks, indicating that their intermediate token representations provide a weaker signal for on-the-fly fault detection, causing the rejection mechanism to retain lower-quality partial sequences longer than necessary. Phi-3-mini occupies a middle ground, offering competitive token counts on GSM8K (7.485M) and MMLU-Chem (0.229M) with broadly stable accuracy, making it a practical alternative under tighter resource constraints.

\subsection{Comparison with existing early-exit methods.}
Table~\ref{tab:method_comparison_minimal} situates our approach relative to existing
early-exit methods \textsc{Deer} and \textsc{LYNX} on \textsc{GSM8K}. Existing early-exit methods operate on synthetic reasoning traces
and reduce token usage by terminating generation at heuristically determined exit
points; however, they do not monitor the \emph{quality} of the generation on the fly
and therefore cannot catch faulty reasoning as it unfolds.

\begin{table}[!ht]
\centering
\scriptsize
\caption{Comparison of early-discard methods on GSM8K. Traditional = no early discard. DEER / LYNX = single-stage discard. Our Method + LYNX = proposed multi-stage early discard. Lower Total Token ($\downarrow$) and higher Throughput ($\uparrow$) and Eval Accuracy ($\uparrow$) are better. Best per column in \textbf{bold} (excluding Traditional baseline).}
\label{tab:method_comparison_minimal}
\begin{tabular}{l c c c}
\toprule
\textbf{Method} & \textbf{Total Token} $\downarrow$ & \textbf{Throughput} $\uparrow$ & \textbf{Eval Accuracy} $\uparrow$ \\
\midrule
Traditional & 109.091M & 3,291 & 0.720 \\
\midrule
DEER & 26.709M & 8,086 & 0.671 \\
LYNX & 25.143M & 8,588 & 0.738 \\
\midrule
MSFIR & 24.611M & 15,823 & 0.734 \\
MSIFR + LYNX & 23.731M & 16,605 & 0.735 \\
\bottomrule
\end{tabular}
\end{table}

The baseline consumes 109.09M tokens at a throughput of 3,291~tokens/s
with an evaluation accuracy of 0.720. \textsc{DEER} and \textsc{LYNX}
reduce token usage to 26.71M and 25.14M respectively by applying early-exit
strategies, yet \textsc{DEER} incurs a 4.9 point accuracy degradation (0.671).
\textsc{LYNX} yields 0.738, marginally above the traditional baseline, and
achieves a modest throughput gain to 8,588~tokens/s but provides no mechanism for
detecting faulty reasoning mid-generation.

\textsc{MSFIR}, applied without any early-exit component, reduces token
consumption to 24.611M and raises throughput to 15,823~tokens/s a
\textbf{77.4\% token reduction} and \textbf{4.8$\times$ throughput improvement}
over the traditional baseline, while maintaining an accuracy of 0.734,
competitive with \textsc{LYNX} (0.738). This demonstrates that in-flight
rejection alone captures nearly all available efficiency gains.

Augmenting \textsc{LYNX} with our multi-stage in-flight rejection achieves the best
result across all three dimensions simultaneously. Total token consumption is further
reduced to 23.73M a \textbf{78.2\% reduction} over traditional generation and
throughput reaches 16,605~tokens/s, a \textbf{5.0$\times$ improvement} over
traditional generation and approximately \textbf{2$\times$} over either early-exit
method used alone. The incremental gain of \textsc{MSIFR + LYNX} over
\textsc{MSFIR} (23.73M vs.\ 24.61M tokens; 16,605 vs.\ 15,823~tokens/s)
confirms that the two mechanisms operate on orthogonal cost axes and compound
cleanly. Evaluation accuracy (0.735) remains on par with \textsc{LYNX} (0.738),
confirming that the additional token savings introduce no accuracy penalty.

Taken together, these results establish that multi-stage in-flight rejection is
orthogonal and complementary to existing early-exit methods: early-exit strategies
reduce unnecessary computation at the end of generation, while our method eliminates
wasted tokens from faulty sample during generation, and combining both
yields compounding efficiency gains that neither approach achieves alone.

\section{Limitations and Conclusion}
\textbf{Limitations.} While MSIFR's rule-based validators are inherently task-specific by design, each validator stage encodes structural and logical constraints native to its target domain, and adapting to a new task requires defining analogous domain-appropriate rejection criteria rather than retraining any model component. While we provide ablation studies on the mid-solution cutoff across all seven benchmarks (Appendix \ref{msc}), showing 50\% to be consistently optimal, domains with substantially different solution length distributions such as very long-form proofs or code generation may require retuning this threshold. Additionally, all experiments are conducted on 7-8B parameter models, leaving behavior at larger scales as an open direction. \\
\textbf{Conclusion.} We presented Multi-Stage In-Flight Rejection (MSIFR), a framework that intercepts and discards low-quality synthetic data generation trajectories, reducing compute wastage. We formalized this procedure as a decision-theoretic optimization problem, showing that any non-trivial discard policy strictly reduces expected token consumption under mild assumptions and establishing a martingale property that guarantees intermediate stopping decisions are free of systematic bias. Empirically, MSIFR delivers consistent gains across multiple models and benchmarks, reducing token consumption by up to 78\% while preserving or improving evaluation accuracy in the majority of model–benchmark pairs. When composed with LYNX, MSIFR achieves a 78.2\% total token reduction and a $5\times$ throughput improvement over conventional generation with no measurable accuracy penalty, demonstrating its ability to compound with existing state-of-the-art. 

\newpage
\bibliography{ref.bib}
\bibliographystyle{plainnat}
\newpage
\section{Appendix}
\subsection{Ablation Study}
\subsubsection{Selection of Mid-Solution Cutoff}
\label{msc}
To determine the optimal mid-solution cutoff, we ablate the threshold from 30\% to 80\%, 
reporting training token count, throughput, and GSM8K accuracy in Table~\ref{tab:ablation_midsol}. Increasing the cutoff monotonically reduces training tokens, from 11.4M at 30\% to 8.1M at 
80\%. However, GSM8K accuracy peaks at 0.57 for cutoffs between 30\%--50\% and drops back 
to the baseline level of 0.56 beyond that, indicating that aggressive truncation discards 
steps with meaningful learning signal. The \textbf{50\% cutoff} is the elbow point: it 
matches peak accuracy while reducing total tokens by 21.6\% relative to the Traditional 
baseline, with no throughput degradation. Beyond 50\%, further token reduction yields no 
accuracy gain. We therefore adopt 50\% as the default configuration for GSM8K. We similarly determined the optimal cutoff points for the other benchmarks and found that a 50\% threshold consistently yielded the best performance across all of them. Due to computational constraints, results are reported without variance across multiple seeds; however, all experiments use a fixed seed for consistency.

\begin{table}[!ht]
\centering
\scriptsize
\caption{Ablation study on mid-solution cutoff percentage. (*) denotes our selected configuration.}
\label{tab:ablation_midsol}
\begin{tabular}{lrrr}
\toprule
\textbf{Mid-Sol. \%} & \textbf{Total Tokens} & \textbf{Throughput (pairs/hr)} & \textbf{GSM8K} \\
\midrule
30\%          & 11,427,523 & 32,137 & 0.57 \\
40\%          & 10,121,512 & 33,049 & 0.57 \\
\textbf{50\% }(*)      &  \textbf{8,951,127} & \textbf{32,141} & \textbf{0.57} \\
60\%          &  8,368,513 & 32,114 & 0.56 \\
80\%          &  8,113,110 & 31,914 & 0.56 \\
\bottomrule
\end{tabular}
\end{table}

\subsubsection{False-Positive and False-Negative Analysis}

To assess whether early rejection discards valuable data (false positives) or fails to intercept bad trajectories (false negatives), we conduct an oracle experiment on a 1,000-sample subset of each benchmark using Llama-3.1-8B.

\paragraph{Setup.}
For each prompt, we first generate a full solution without early stopping and label it as \textit{Good} (passes all final validators) or \textit{Bad} (fails at least one). We then replay the partial outputs through MSIFR's stage-wise validators without terminating, recording whether the trajectory would have been rejected early. A \textit{false positive} occurs when a Good trajectory is rejected at any intermediate stage. A \textit{false negative} occurs when a Bad trajectory survives all intermediate checks and is only rejected at final evaluation.

\paragraph{Results.}
Table~\ref{tab:ablate_fp_fn} shows that the false positive rate remains below \(5\%\) across all benchmarks, indicating minimal loss of high-quality training data. The false negative rate averages \(8.7\%\); these trajectories still consume fewer tokens than traditional full generation because final rejection occurs before completion.

\begin{table}[h]
\centering
\scriptsize
\caption{False-positive and false-negative analysis (Llama-3.1-8B, 1,000 samples per benchmark).}
\label{tab:ablate_fp_fn}
\begin{tabular}{lcccc}
\toprule
\textbf{Benchmark} & \textbf{\#Good} & \textbf{\#Bad} & \textbf{FPR (\%)} & \textbf{FNR (\%)} \\
\midrule
GSM8K             & 742 & 258 & \(3.1\) & \(7.4\) \\
MATH500           & 621 & 379 & \(4.2\) & \(9.0\) \\
MMLU-Chem         & 573 & 427 & \(2.4\) & \(7.3\) \\
SVAMP             & 687 & 313 & \(2.8\) & \(6.4\) \\
MAWPS             & 721 & 279 & \(2.9\) & \(6.5\) \\
MathQA            & 802 & 198 & \(3.5\) & \(10.6\) \\
DeepMind Math     & 754 & 246 & \(3.6\) & \(8.9\) \\
\midrule
\textbf{Average}  & --  & --  & \(\mathbf{3.2}\) & \(\mathbf{8.7}\) \\
\bottomrule
\end{tabular}
\end{table}

Table~\ref{tab:ablate_sensitivity} shows the trade-off as the mid-solution cutoff varies on GSM8K. The default \(50\%\) cutoff balances low FPR (\(3.1\%\)) against strong token savings (\(42\%\)).

\begin{table}[!ht]
\centering
\scriptsize
\caption{Sensitivity of FP/FN rates to mid-solution cutoff (GSM8K).}
\label{tab:ablate_sensitivity}
\begin{tabular}{lccc}
\toprule
\textbf{Cutoff} & \textbf{FPR (\%)} & \textbf{FNR (\%)} & \textbf{Token Savings (\%)} \\
\midrule
30\%  & \(1.2\) & \(14.5\) & \(22\%\) \\
40\%  & \(2.0\) & \(11.2\) & \(31\%\) \\
\textbf{50\%} & \(\mathbf{3.1}\) & \(\mathbf{7.4}\) & \(\mathbf{42\%}\) \\
60\%  & \(5.8\) & \(4.1\) & \(51\%\) \\
80\%  & \(11.3\) & \(2.0\) & \(58\%\) \\
\bottomrule
\end{tabular}
\end{table}

\subsection{Chat Template}
\label{app:chat_template}

To ensure correct tokenization and prompt formatting for each instruction-tuned model, we apply the model-specific chat template as defined by each model's official release.
Table~\ref{tab:enhanced_templates} summarizes the chat templates used for all five models evaluated in this work.
Models follow three distinct formatting conventions: ChatML (Qwen2.5-7B-Instruct), Llama-3 header-based format (Llama-3.1-8B-Instruct), simplified tag-based format (Phi-3-Mini-Instruct), bracket-based instruct format (Mistral-7B-Instruct), and a plain-text user--assistant format (DeepSeek-7B-Chat).
Applying the correct chat template is critical for in-flight rejection, as malformed prompts or incorrect role tokens can corrupt the partial generation signal used by our multi-stage validator, leading to spurious rejections or missed faults.
All templates are applied consistently across training, validation, and inference stages.

\begin{table*}[!ht]
\centering
\small
\setlength{\tabcolsep}{3pt}
\caption{Chat Templates for Popular Instruction-Tuned LLMs}
\label{tab:enhanced_templates}
\rowcolors{1}{white}{lightgray!10}
\begin{tabularx}{\linewidth}{@{} >{\bfseries}l >{\ttfamily\small}l X @{}}
\toprule
\rowcolor{teal!20}
\textbf{Model} & \textbf{Format Type} & \textbf{Template Pattern} \\
\midrule
Qwen2.5-7B-Instruct & ChatML & <|im\_start|>role\textbackslash ncontent<|im\_end|>\textbackslash n \\
\cmidrule{2-3}
& System & <|im\_start|>system\textbackslash n\{system\}<|im\_end|>\textbackslash n \\
& User & <|im\_start|>user\textbackslash n\{user\}<|im\_end|>\textbackslash n \\
& Assistant & <|im\_start|>assistant\textbackslash n \\
\midrule
Llama-3.1-8B-Instruct & Llama3 & <|start\_header\_id|>role<|end\_header\_id|>\textbackslash n\textbackslash ncontent<|eot\_id|> \\
\cmidrule{2-3}
& System & <|start\_header\_id|>system<|end\_header\_id|>\textbackslash n\textbackslash n\{system\}<|eot\_id|> \\
& User & <|start\_header\_id|>user<|end\_header\_id|>\textbackslash n\textbackslash n\{user\}<|eot\_id|> \\
& Assistant & <|start\_header\_id|>assistant<|end\_header\_id|>\textbackslash n\textbackslash n \\
& Prefix & <|begin\_of\_text|> \\
\midrule
DeepSeek-7B-Chat & Simple & User: \{user\}\textbackslash n\textbackslash nAssistant: \\
\midrule
Phi-3-Mini-Instruct & Simplified & <|role|>\textbackslash ncontent<|end|>\textbackslash n \\
\cmidrule{2-3}
& User & <|user|>\textbackslash n\{user\}<|end|>\textbackslash n \\
& Assistant & <|assistant|>\textbackslash n \\
\midrule
Mistral-7B-Instruct & Instruct & [INST] \{user\} [/INST] \\
\bottomrule
\end{tabularx}
\end{table*}

\subsection{Validator}
\label{app:validator}

Our Early Detection Validator (EDV) is a lightweight, rule-based three-stage framework designed to intercept faulty generations at the earliest possible token position without relying on a secondary neural model.
Table~\ref{tab:validation_framework} details the operational description and rejection criteria for each stage.

The first stage, Well-Posedness Enforcer (WPE), executes at 0\% generation progress and validates the structural integrity of the input prompt itself rejecting ill-formed inputs before any generation budget is spent.
The second stage, Reasoning Trace Auditor (RTA), activates at the 50\% generation checkpoint and audits the partial reasoning trace for arithmetic consistency, hallucination signals, and premature termination markers, enabling the system to abort generations that have already diverged from a sound reasoning trajectory at the midpoint rather than waiting for full completion.
The third stage, Solution Convergence Validator (SCV), applies at generation completion (100\%) and enforces final answer alignment, formatting closure, and overflow constraints, providing a last-resort filter for generations that passed earlier stages but fail to produce a well-formed solution.

Together, the three stages implement a progressive rejection cascade in which each successive stage handles faults that are only detectable with more generation context, ensuring that the rejection decision is made at the earliest token position at which it is reliably diagnosable.
Beyond rule-based validation, a post-hoc Verification Suite evaluates accepted generations using LLM-as-Judge scoring, MinHash-based duplication detection, and human judgment on a 100-sample subset, providing an empirical upper bound on the quality of generations that pass all three EDV stages.

\begin{table}[!ht]
\centering
\small
\caption{Three-Stage Progressive Validation Framework for Early Detection Validator (EDV) with Evaluation Protocol. Component abbreviations: WPE = Well-Posedness Enforcer, RTA = Reasoning Trace Auditor, SCV = Solution Convergence Validator.}
\label{tab:validation_framework}
\begin{tabular}{@{}cp{2.2cm}p{4.8cm}p{5.5cm}@{}}
\toprule
\textbf{Stage} & \textbf{Component} & \textbf{Operational Description} & \textbf{Rejection Criteria} \\
\midrule
$S_1$ & Well-Posedness Enforcer & Validates input well-posedness: punctuation (?), numbers, length (8--100 words), English-only. & \textbf{Rejects} ill-formed prompts: missing ?; non-English (CJK, Cyrillic, Arabic); <8 or >100 words; cut-off suffix; repeated punctuation; lowercase start. \\
\midrule
$S_2$ & Reasoning Trace Auditor & Audits partial reasoning: arithmetic soundness, early-exit suppression (premature \#\#\#\# with text), duplicate markers, magnitude sanity. & \textbf{Rejects} hallucinated/ dissonant traces: hallucination phrases; premature \#\#\#\# with text; duplicate \#\#\#\#; arithmetic error; magnitude > 100× problem max or > 10M (absolute cap); unexpected negatives. \\
\midrule
$S_3$ & Solution Convergence Validator & Enforces final answer alignment, formatting closure, trace consistency, overflow prevention. & \textbf{Rejects} non-convergent solutions: final answer mismatch; missing markers; cut-off end; prompt leakage (\texttt{SYSTEM:}, \texttt{User:}); duplicate finals; magnitude >10M (The absolute 10M cap ensures consistency between stages; problems with max values above 10M are outside the intended benchmark distribution.). \\
\midrule
Eval & Verification Suite & Evaluate with LLM-as-Judge, MinHash duplication detection, Human judgment (N=100 samples). & \textbf{Rejects} solutions with LLM rating < 3 (on 1-5 scale); rejects any duplicate detected via MinHash; rejects solutions marked incorrect by majority of human judges (N=100). \\
\bottomrule
\end{tabular}
\end{table}

While the specific validation rules are task-dependent, the framework is modular: each stage can be adapted by defining domain-specific constraints without modifying the underlying model. This allows MSIFR to generalize across tasks by replacing validators rather than retraining components.

\subsection{LLM Usage}

We used a large language model (LLM) solely for writing polish, such as improving grammar, clarity, and readability of the text. The LLM did not contribute to research ideation, methodology, analysis, or results. All scientific content and conclusions are the responsibility of the authors.
\newpage
\begin{figure}[H]
\centering
\includegraphics[width=0.97\textwidth]{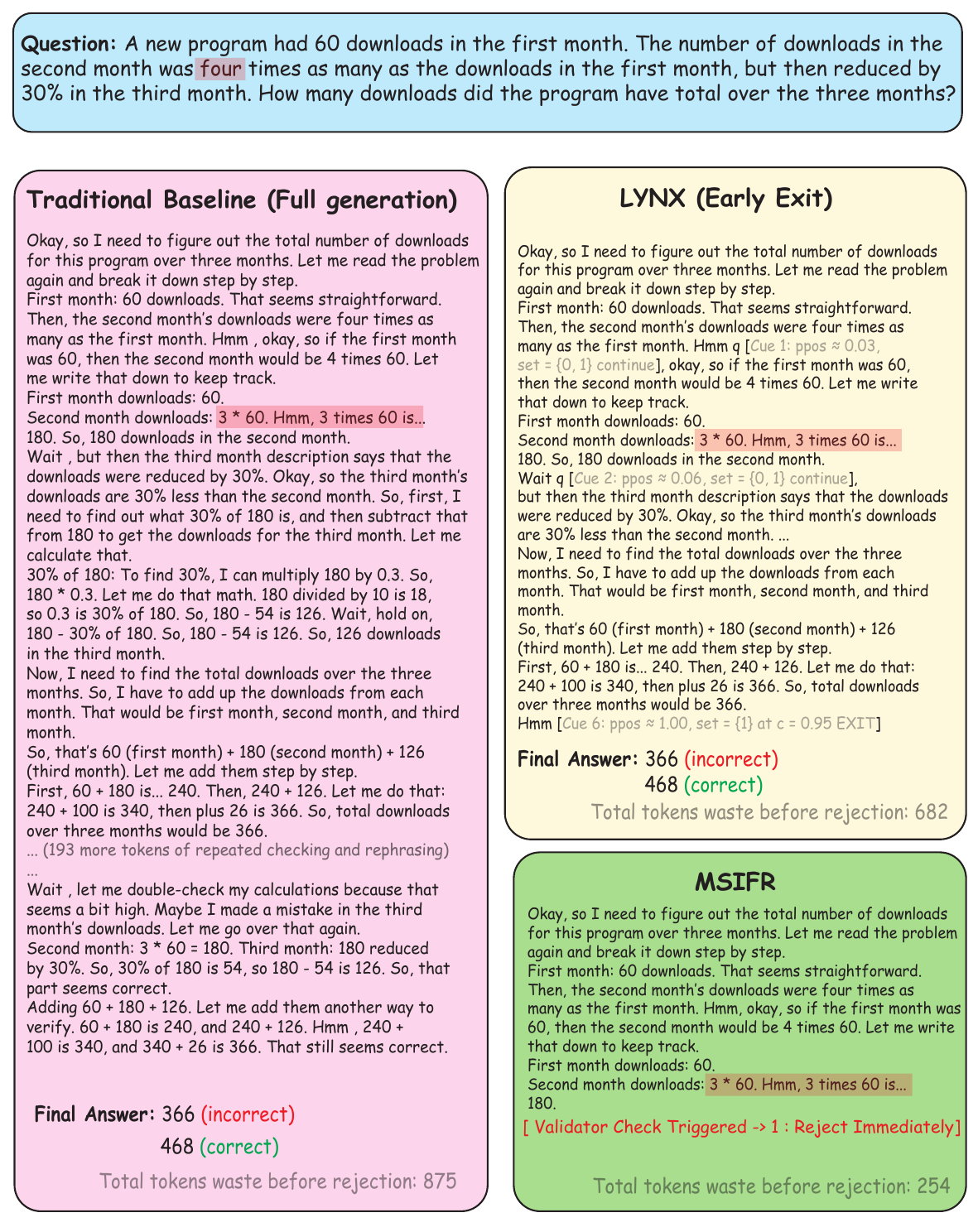}
\caption{Comparison of three approaches to handling faulty arithmetic reasoning in synthetic data generation. 
The Traditional baseline generates the full reasoning trace before applying validation, resulting in 
875 tokens wasted on an incorrect solution. In this example, the model incorrectly computes the second 
month as $3 \times 60$ instead of $4 \times 60$, leading to an incorrect final answer (366 instead of 
the correct answer, 468). Because validation occurs only after full generation, the error is detected 
only after all tokens are consumed.
LYNX improves efficiency by introducing probabilistic early-exit cues, reducing wasted tokens to 682. 
However, its stopping decision is triggered only after sufficient confidence is accumulated, which 
typically occurs late in the generation process, allowing incorrect reasoning to progress substantially 
before termination.
MSIFR performs in-flight rejection by applying validation checks during generation. Once the arithmetic 
inconsistency is detected at the intermediate stage, generation is immediately terminated at 254 tokens, 
yielding a 71\% reduction in wasted tokens relative to the Traditional baseline. By intercepting faulty 
trajectories early, MSIFR reduces unnecessary token consumption and prevents propagation of incorrect 
reasoning into the retained dataset.}
\label{fig2}
\end{figure}

\end{document}